\documentclass[10pt,twocolumn]{article}

\usepackage{iccv}
\usepackage{times}
\usepackage{epsfig}
\usepackage{graphicx}
\usepackage{amsmath}
\usepackage{amssymb}
\usepackage{epstopdf}
\usepackage{booktabs}
\usepackage{appendix}
\usepackage{multirow}
\usepackage{rotating}
\usepackage{xr}
\usepackage[table,xcdraw]{xcolor}
\usepackage[normalem]{ulem}
\usepackage{amssymb,mathrsfs,amsmath}
\usepackage[ruled,vlined,linesnumbered]{algorithm2e}
\useunder{\uline}{\ul}{}

\usepackage[pagebackref=true,breaklinks=true,colorlinks,bookmarks=false]{hyperref}

\iccvfinalcopy % *** Uncomment this line for the final submission

\begin{document}

%%%%%%%%% TITLE
\title{SaliencyCut: Augmenting Plausible Anomalies for Anomaly Detection}

% \author{Jianan Ye\\
% \and

\author{Jianan Ye$^{1,2,\ast}$, Yijie Hu$^{1,2,}$\thanks{Equal contribution.}, Xi Yang$^{1}$, Qiu-Feng Wang$^{1}$, Chao Huang$^{3}$, Kaizhu Huang$^{4,}$\thanks{Corresponding author.}\\
$^{1}$School of Advanced Technology, Xi’an Jiaotong-Liverpool University\\
$^{2}$Department of Electrical Engineering and Electronics, University of Liverpool\\
$^{3}$Department of Computer Science, University of Liverpool\\
$^{4}$Data Science Research Center, Duke Kunshan University\\
\tt\small \{Jianan.Ye20, Yijie.Hu20\}@student.xjtlu.edu.cn, \{Xi.Yang01, Qiufeng.Wang\}@xjtlu.edu.cn,
\\\tt\small Chao.Huang2@liverpool.ac.uk, kaizhu.huang@dukekunshan.edu.cn
}

\maketitle

%%%%%%%%% ABSTRACT
\begin{abstract}
   Anomaly detection under open-set scenario is a challenging task that requires learning discriminative fine-grained features to detect anomalies that were even unseen during training. As a cheap yet effective approach, data augmentation has been widely used to create pseudo anomalies for better training of such models. Recent wisdom of augmentation methods focuses on generating random pseudo instances that may lead to a mixture of augmented instances with seen anomalies, or out of the typical range of anomalies. To address this issue, we propose a novel saliency-guided data augmentation method, SaliencyCut, to produce pseudo but more common anomalies which tend to stay in the plausible range of anomalies. Furthermore, we deploy a two-head learning strategy consisting of normal and anomaly learning heads, to learn the anomaly score of each sample. Theoretical analyses show that this mechanism offers a more tractable and tighter lower bound of the data log-likelihood. We then design a novel patch-wise residual module in the anomaly learning head to extract and assess the fine-grained anomaly features from each sample, facilitating the learning of discriminative representations of anomaly instances. Extensive experiments conducted on six real-world anomaly detection datasets demonstrate the superiority of our method to competing methods under various settings.
\end{abstract}

%%%%%%%%% BODY TEXT
\section{Introduction}

Anomaly detection is designed to detect defective samples that exhibit minor deviations from expected data patterns yet share similarities with the primary patterns of normal samples~\cite{li2021cutpaste,ding2022catching,pang2019deep,pang2021explainable}.
Unlike conventional open-set or out-of-distribution methods~\cite{liu2018open,vaze2022generalized,bendale2016towards,zaeemzadeh2021out,sun2022out}, open-set fine-grained anomaly detection targets both seen (\emph{e.g.}, color spots on leather surfaces) and unseen types of anomalies (\emph{e.g.}, folds or glues on leather surfaces) within a single object class during training. 
%-------------------------insert banner---------------------------------------------%
\begin{figure}[t]
\centering
\includegraphics[scale=0.45]{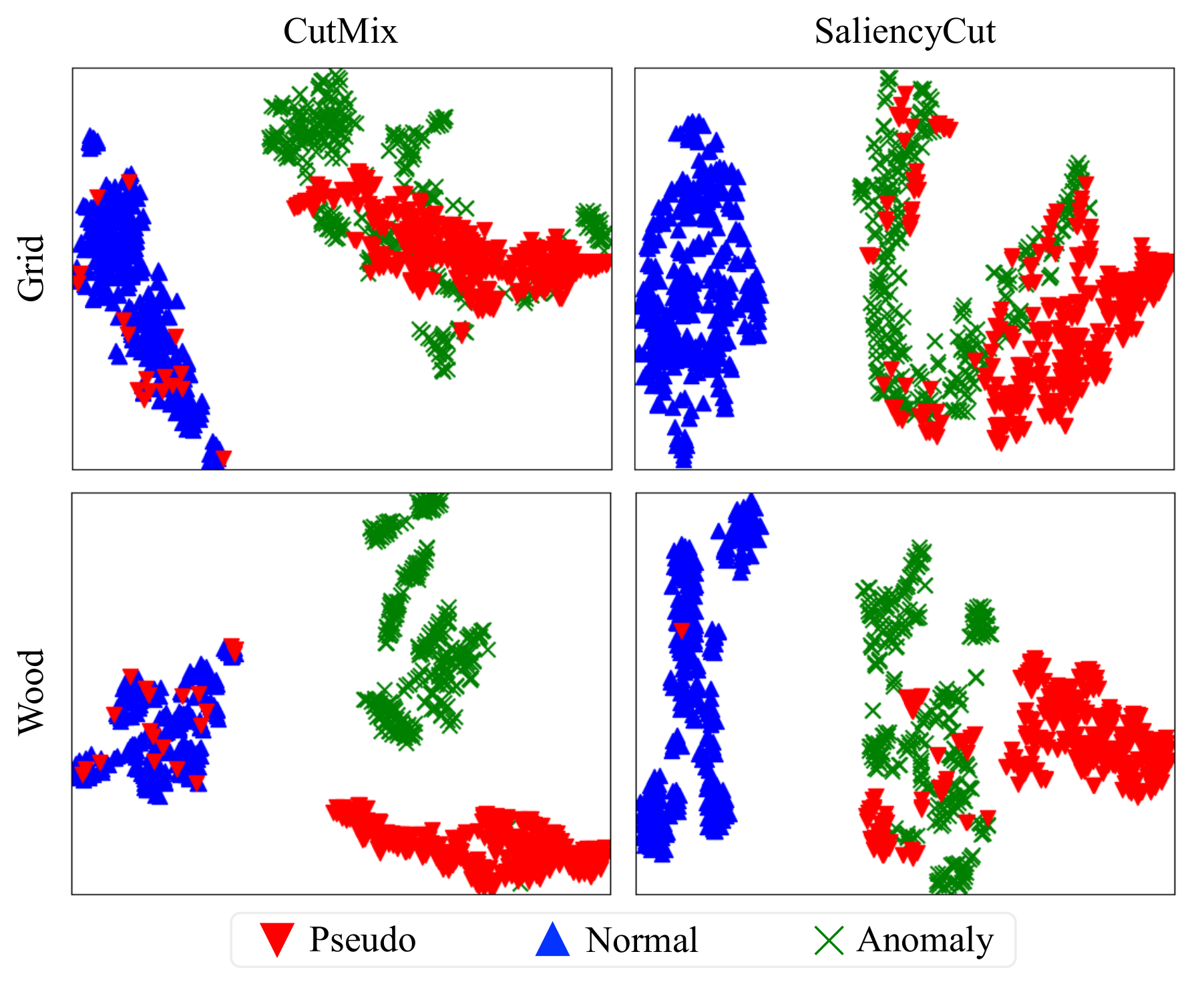}
\caption{
t-SNE visualizations of data distribution in the feature space during training on two datasets. Pseudo anomaly samples are created by CutMix~\cite{yun2019cutmix} in a random way which tend to be mixed with both seen anomalous and normal samples, or overly expand the training anomaly samples (red triangles in left-bottom). Our method migrates these issues. }
\label{fig:banner}
\end{figure}
%-------------------------insert banner---------------------------------------------%
To deal with this open-set problem, it is crucial for the detection model to have the ability to identify both seen and unseen types of anomalies. Retraining the model every time new anomaly data appears is a straightforward solution, but it can be inefficient and costly.
Additionally, even for seen anomalies,  there may be a scarcity of available samples, which makes this problem even more challenging.

Given that anomalies are akin to normal samples, adding perturbation is one cheapest way to expand anomalies distribution for tackling open-set fine-grained anomaly detection. Current efforts~\cite{ding2022catching, li2021cutpaste} focus on creating pseudo instances by data augmentation methods like CutMix~\cite{yun2019cutmix} or CutPaste~\cite{li2021cutpaste}.
By simulating the possible types of unseen instances, more diverse pseudo anomalies are involved during training to learn rich discriminative anomaly features. However, lacking the guidance of real anomaly information, such methods usually synthesize pseudo instances in a random way. As a result, the pseudo anomalies could either overlap with the distribution of the seen training instances or unduly expand the distribution of the anomalies. Figure~\ref{fig:banner} illustrates one simple example. In the left part, Cutmix, one typical data augmentation method generates pseudo anomaly instances that are mixed with both seen normal and anomaly instances in Grid data (left-top of Figure~\ref{fig:banner}). For the Wood data (left-bottom of Figure~\ref{fig:banner}), it is evident that the pseudo instances unduly expand the distribution of the anomalies. Obviously, pseudo anomalies either mixed with the existing samples or staying too farther away from anomalies may confuse the  detection model and thus limit the performance.

\begin{figure}[t]
    \centering
    \includegraphics[scale=0.48]{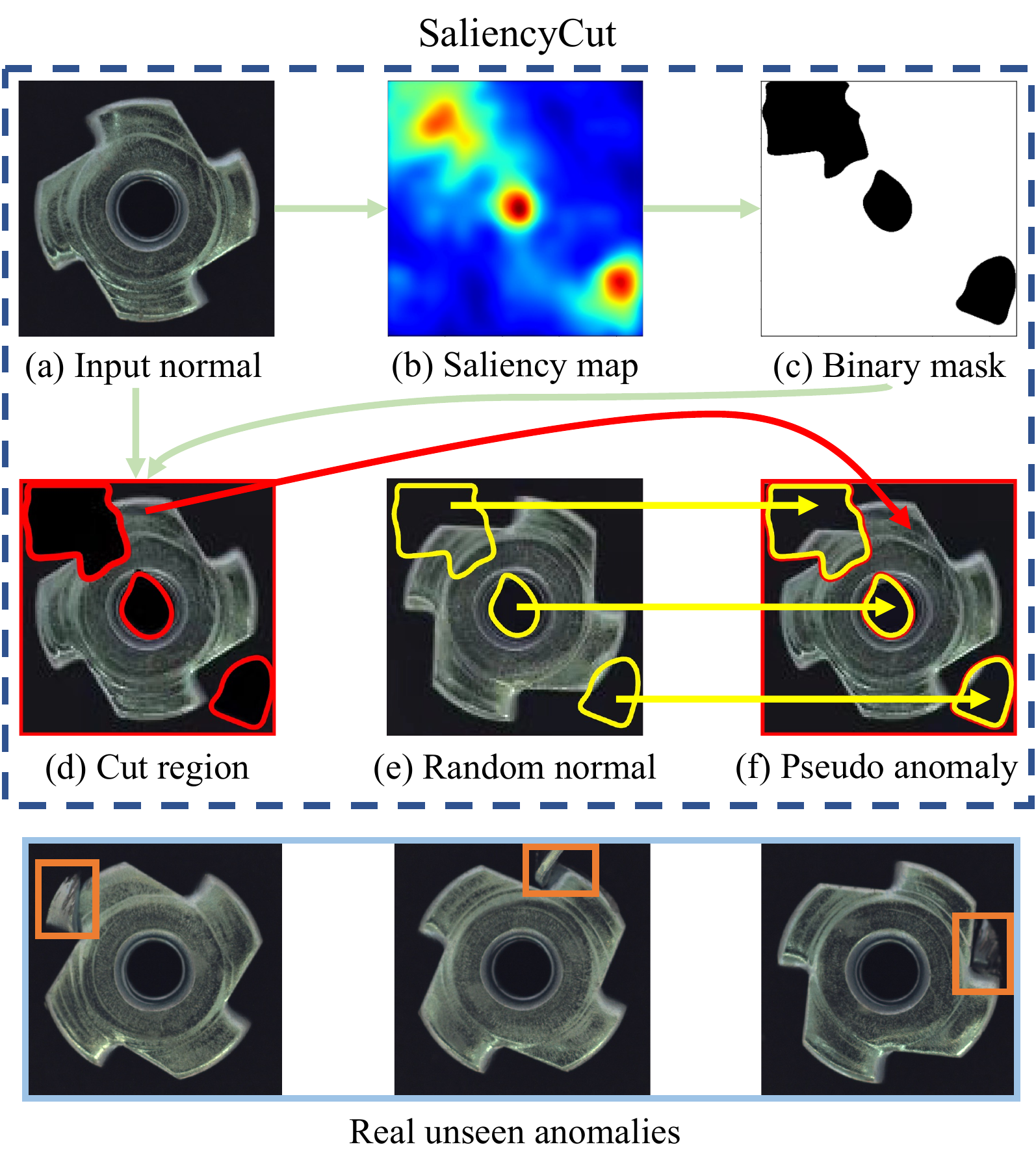}
    \caption{Illustration of SaliencyCut for generating pseudo anomalies. Real unseen anomalies are shown in the last low, anomaly regions are marked with orange boxes. The resulting pseudo anomaly looks very similar to real unseen anomalies.}
    \label{fig:saliencycut}
\end{figure}

In this paper, we argue that one ideal anomaly-data augmentation method should generate samples that are neither far away nor mixed with seen anomalies. Moreover, it is observed that  salient regions of normal instances representing the typical normal regions would be essentially more likely to be recognized as normal instances instead of anomalies, while non-salient regions could have a higher chance to contain  anomaly information.  Motivated by these observations,
we present a novel saliency-guided data augmentation method called \textbf{\emph{SaliencyCut}} to mitigate the defects of random pseudo-anomalies generation methods (\eg, CutMix or CutPaste), \ie, we aim to generate pseudo anomaly data guided by saliency so that they would not be mixed with normal instances but tend to stay in the plausible range of anomalies.
To do so, 1) we first obtain the saliency map of a normal instance by calculating the backward propagated gradients, thus the saliency part would stand for more typical or reliable normal instance regions; 2) we then cut  the non-salient regions from the normal instance, and paste them  to any other random normal instance to obtain a final pseudo anomaly. 
See Figure~\ref{fig:saliencycut} for the detailed  flowchart of our Saliencycut. While the saliency region
generally indicates the more reliable or confident part as a normal instance, the non-salient parts would tend to stay farther from a normal instance. Namely, they have a higher chance to distribute closer to anomalies.
By mixing such non-salient regions, the resulting pseudo instances tend to stay closer to the anomalies and would be less likely to overlap with normal instances. 
As clearly illustrated in Figure~\ref{fig:banner}, in the right part, pseudo instances generated by our SaliencyCut  are indeed rarely mixed with normal instances and also expand the plausible region of anomalies. As also observed in the last two rows of Figure~\ref{fig:saliencycut}, the resulting augmented  sample (f) looks much similar to the real anomalies (last row).

On the implementation front, in order to  enable the model to further discriminate against all types of anomalies with SaliencyCut, we design a novel two-head learning mechanism to learn the score distribution of normal and anomalous samples. One head focuses on learning the features of normal samples,
while the other focuses on learning fine-grained anomaly features through anomalous residuals within extracted features,
thus promoting the model to discriminate against anomalous samples.
To efficiently learn such anomalous residuals,
we propose one patch-wise residual module, where residual features within each sample are calculated for enlarging the gap between normal and anomaly samples.
As a result, our model learns abnormality representations, which can discriminate both seen and unseen anomalies from the normal data, as shown in Figure~\ref{fig:dtsne}.
As an important contribution, we have provided theoretical analysis showing that our two-head learning mechanism offers on the data log-likelihood  a tighter lower bound than the four-head approach~\cite{ding2022catching} that is more tractable to optimize, thus justifying our novel framework. 

We summarize our contributions as follows: 1) We propose a saliency-guided data augmentation, namely SaliencyCut, creating pseudo anomalies to expand the range of anomaly samples, improving the model's detection ability on unseen anomalies. 2) We deploy a two-head learning mechanism based on the theoretical lower bound in the anomaly score learning process. 3) We further design a novel patch-wise residual module to gather fine-grained anomaly knowledge. 4)  We conduct comprehensive experiments on six real application datasets. The results show that our model substantially outperforms competing methods under various settings.

\section{Related Work}
\label{sec:relatedwork}
{\bf Open-set/Out-of-Distribution Detection.} Open-set detection and out-of-distribution detection are recognized as coarser-grained detection methods. Open-set detection~\cite{liu2018open,vaze2022generalized,bendale2016towards,Zhou2021LearningPF,Yue2021CounterfactualZA} seeks to identify novel object classes that were not accounted for during training, and out-of-distribution detection~\cite{Huang2021MOSTS,lin2021mood,zaeemzadeh2021out,sun2022out} seeks to detect data that is significantly dissimilar from the training data. Our task, on the other hand, is oriented towards fine-grained detection, with the aim of identifying subtle anomaly characteristics in samples of a single object within an open-set scenario.

{\bf Supervised Anomaly Detection Methods.} In recent years, supervised anomaly detection attracts increasing attention, which aims to enhance the ability of deep learning models to identify anomaly samples by learning anomaly features from a small number of seen anomalies. One-class metric learning framework~\cite{gornitz2013toward,liu2019margin,pang2018learning,ruff2019deep} with anomalies as negative samples or models learning with deviation loss~\cite{pang2019deep,zhang2020viral,pang2021explainable} are proposed to tackle this issue.
As these methods introduce anomaly samples into the training process, a consequent problem is that the models overfit the seen types of anomalies. 

{\bf Anomaly-Free Detection Methods.} 
By assuming that anomaly samples are not available during the training stage, considerable progress has been made in anomaly-free detection studies under the one-classification framework~\cite{chalapathy2018anomaly,chen2022deep,perera2019learning,ruff2018deep}, self-supervised framework~\cite{bergman2020classification,georgescu2021anomaly,golan2018deep,li2021cutpaste,sohn2020learning,tian2021constrained,wang2019effective}, GAN-based framework~\cite{perera2019ocgan,sabokrou2018adversarially,schlegl2019f,zaheer2020old}, and autoencoder-based framework~\cite{gong2019memorizing,hou2021divide,park2020learning,zhou2017anomaly,zhou2020encoding}. 
Despite the fact that they do not suffer the danger of favoring the observed anomalies, it is challenging to distinguish anomaly samples from normal samples due to a lack of real anomalies.

{\bf Augmentations.}
Mixup~\cite{zhang2017mixup} and CutMix~\cite{yun2019cutmix} are two early augmentation methods that mix samples and labels, where the mixed samples are fed into networks as new data, and their corresponding labels are the mixed labels. Similar to CutMix, CutPaste~\cite{li2021cutpaste} extracts a rectangular image patch of variable sizes from a normal image, rotates or jitters pixels in the path, and pastes it back to an image at a random location.
In the same way as CutPaste, DRA~\cite{ding2022catching} employs CutMix to generate pseudo anomaly samples from normal samples to be trained as a third class of samples in addition to normal and anomaly.
Saliency-based data augmentation techniques such as Saliencymix~\cite{uddin2020saliencymix} and SAGE~\cite{Ma2022SAGESM} use saliency regions to improve generalization for classification tasks by using saliency regions for CutMix~\cite{yun2019cutmix} or overall saliency maximization mixup, but also involve label mixing. Our method, SaliencyCut, expands the anomaly distribution by using non-saliency regions of normal samples without label mixing, where augmented samples are treated as pseudo anomalies for expanding anomaly distribution.

\section{Methodology}
\label{sec:method}
\subsection{Problem Statement}
We denote the training anomaly dataset as $\mathcal{D}=\left\{\mathbf{x}_i\right\}_{i=1}^{n+m}$ which consists of $n$ labeled normal samples $\left\{x_1,x_2...x_n\right\}$ and $m$ labeled anomaly samples$\left\{x_{n+1},x_{n+2}...x_{n+m}\right\}$$(n \gg m)$. Here, $m$ anomalies are from seen anomaly types $\mathcal{S} \subset \mathcal{U}$, where $\mathcal{U}=\left\{c_i\right\}_{i=1}^{|\mathcal{U}|}$ denotes all types of anomalies. The goal is to learn an anomaly scoring network, $\phi(\mathcal{D})\rightarrow \mathbb{R}$, which assigns larger anomaly scores to both sorts of anomalies (seen and unseen). 
We predict $\phi(\mathbf{x};\Theta_{h})$, $h\in \left\{1,2\right\}$, of two heads in the training phase, respectively, while performing $\phi(\mathbf{x};\Theta_{2})-\phi(\mathbf{x};\Theta_{1})$ to obtain anomaly scores in inference.

\subsection{Overview of Our Method}

\begin{figure}
\centering
\includegraphics[scale=0.29]{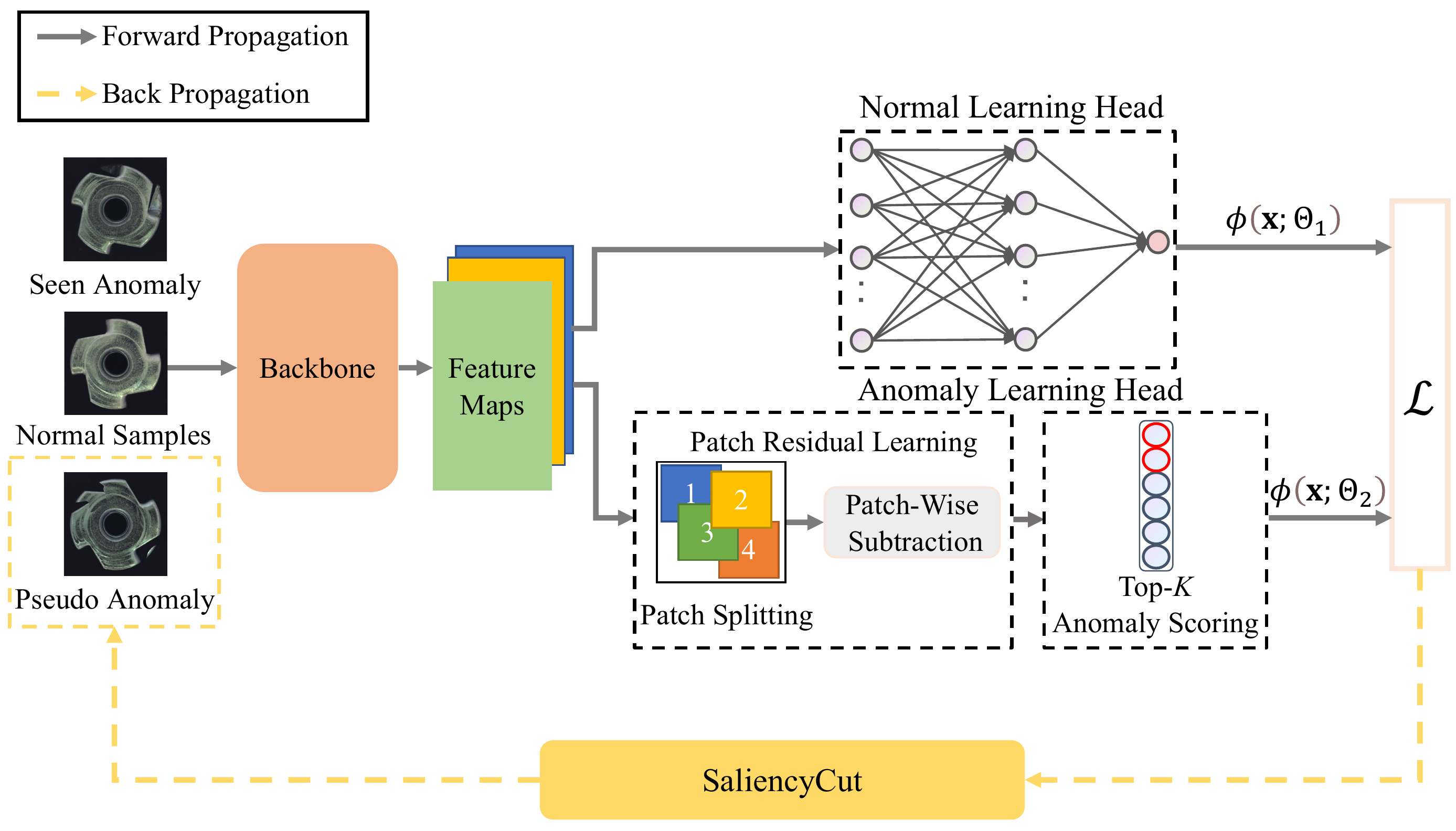}
\hfill
\caption{Overview of our proposed framework.}
\label{fig:network}
\end{figure}

Our proposed method is composed of two major strategies: 1) we develop a data augmentation method that allows for the generation of pseudo but more plausible anomalies that tend to stay within a reasonable range of anomaly data; 2)  we design a novel patch-wise residual module to gather anomaly knowledge and deploy a two-head learning strategy to learn the anomaly score of each sample, which effectively detects both seen and unseen anomalies.

The overall architecture is shown in Figure~\ref{fig:network}. We generate pseudo anomalies with SaliencyCut once before the beginning of each epoch in training, during which parameters of the network are not updated. In this phase, only normal and anomaly samples are fed into the network for forward propagation. The network's gradients are then back-propagated according to the loss function. Saliency regions are calculated by the gradients of each pixel in normal samples, which are deployed by SaliencyCut to generate pseudo anomalies.
Subsequently,  training samples of normal, seen anomalies and pseudo anomalies are fed into the backbone network for feature extraction. Two scoring heads finally determine the anomaly score of each sample based on the extracted features.  Normal Learning Head is constructed by $2$-layers MLP. Anomaly Learning Head consists of a patch-wise residual module (in Section~\ref{method:patch})  and a top-$K$ anomaly scoring module~\cite{pang2019deep, pang2021explainable}.

The network is optimized using the deviation loss ~\cite{pang2019deep, pang2021explainable} in an end-to-end manner, which defines the deviation as a Z-score with Gaussian prior score:
\begin{equation}
\operatorname{dev}\left(\mathbf{x}_i ; \Theta_{h}\right)=\frac{\phi(\mathbf{x}; \Theta_{h})-\mu_{\mathcal{R}}}{\sigma_{\mathcal{R}}}, h\in \left\{1,2\right\},
\end{equation}
where $\mu_{\mathcal{R}}$ and $\sigma_{\mathcal{R}}$ represent the mean and standard deviation of prior-based anomaly score set drown for Gaussian distribution.
Then the deviation loss is specified by:
\begin{equation}
\begin{aligned}
\mathcal{L}=\sum_{h=1,2}[&\left(1-y_i\right)\left|\operatorname{dev}\left(\mathbf{x}_i ; \Theta_{h}\right)\right| \\
&+y_i \max \left(0, a-\operatorname{dev}\left(\mathbf{x}_i ; \Theta_{h}\right)\right)],
\end{aligned}
\end{equation}
where $y = 1$ if $\mathbf{x}_i$ is an anomaly, and $y = 0$ if $\mathbf{x}_i$ is normal. $a$ denotes a Z-score confidence interval parameter, which is set to 5 in our implementation. 
We then provide a comprehensive introduction to each component.

\subsection{SaliencyCut}
\begin{algorithm}[t]
	\caption{SaliencyCut Algorithm}
	\label{alg:algorithm1}
	\KwIn{Normal sample $x_{a}$, detection network $f_\theta$, loss function $\mathcal{L}$.}
	\KwOut{Pseudo Samples $\tilde{\mathbf{x}}$.}  
	\BlankLine
	Calculate gradients of ${Grad}=\nabla_{x_{a}} \mathcal{L}\left(f_\theta\left(x_{a}\right)\right)$;
 
        Saliency map $\boldsymbol{G}_{a}\leftarrow$ normalize $(\text{smooth}(|{Grad}|))$;

        Saliency mask $S_{a} =\text{threshold}(\boldsymbol{G}_{a})$;
        
        Randomly sample another normal instance $x_{b}$;

        Pseudo sample $\tilde{\mathbf{x}} = (1 - S_{a}) \odot {x_{a}} + S_{a} \odot {x_{b}}$;
        \label{alg:odot}

        \textbf{Return} $\tilde{\mathbf{x}}$
\end{algorithm}

\begin{figure}
\centering
\includegraphics[scale=0.3]{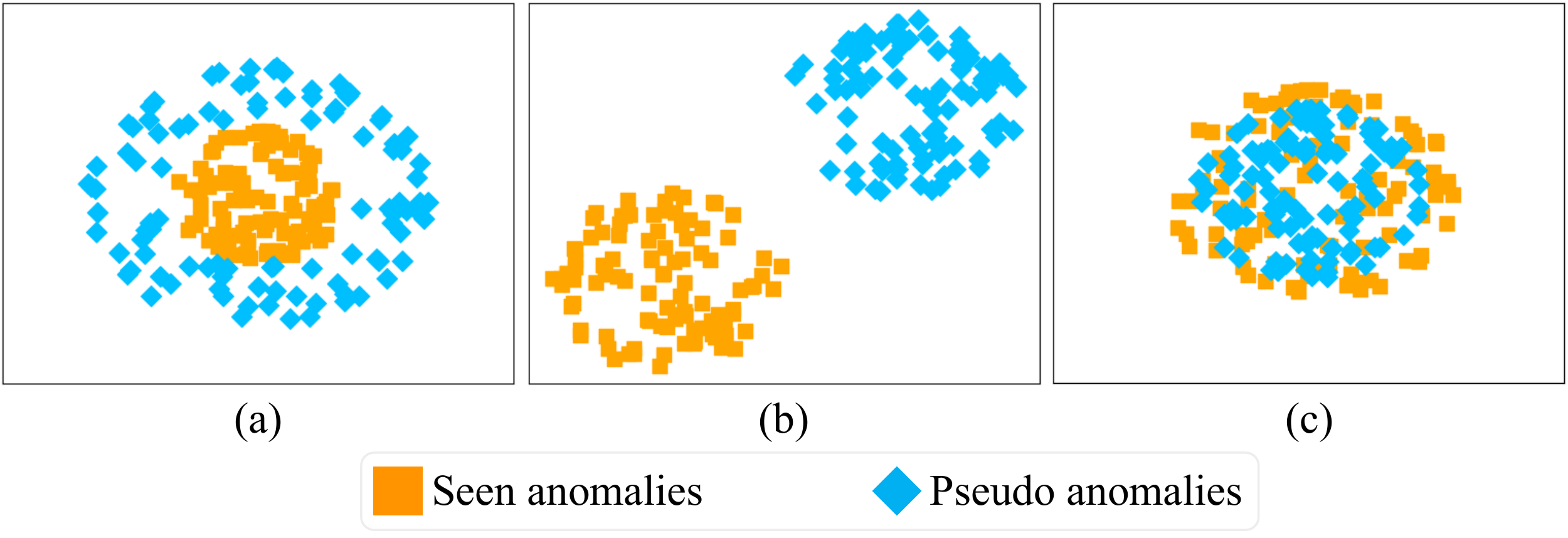}
\caption{Illustration of distributions of seen anomalies and pseudo anomalies. (a) shows one ideal distribution of seen and pseudo anomalies, (b) and (c) are two less effective situations.}
\label{fig:datadis}
\end{figure}

Ideally, the generated pseudo-anomalies are supposed to expand the distribution of known anomalies (\eg, Figure~\ref{fig:datadis}\textcolor{red}{a}) rather than be out of the distributed or mixed with real anomaly samples (\eg, Figure~\ref{fig:datadis}\textcolor{red}{b,c}).

\begin{figure*}
\centering
\includegraphics[scale=0.69]{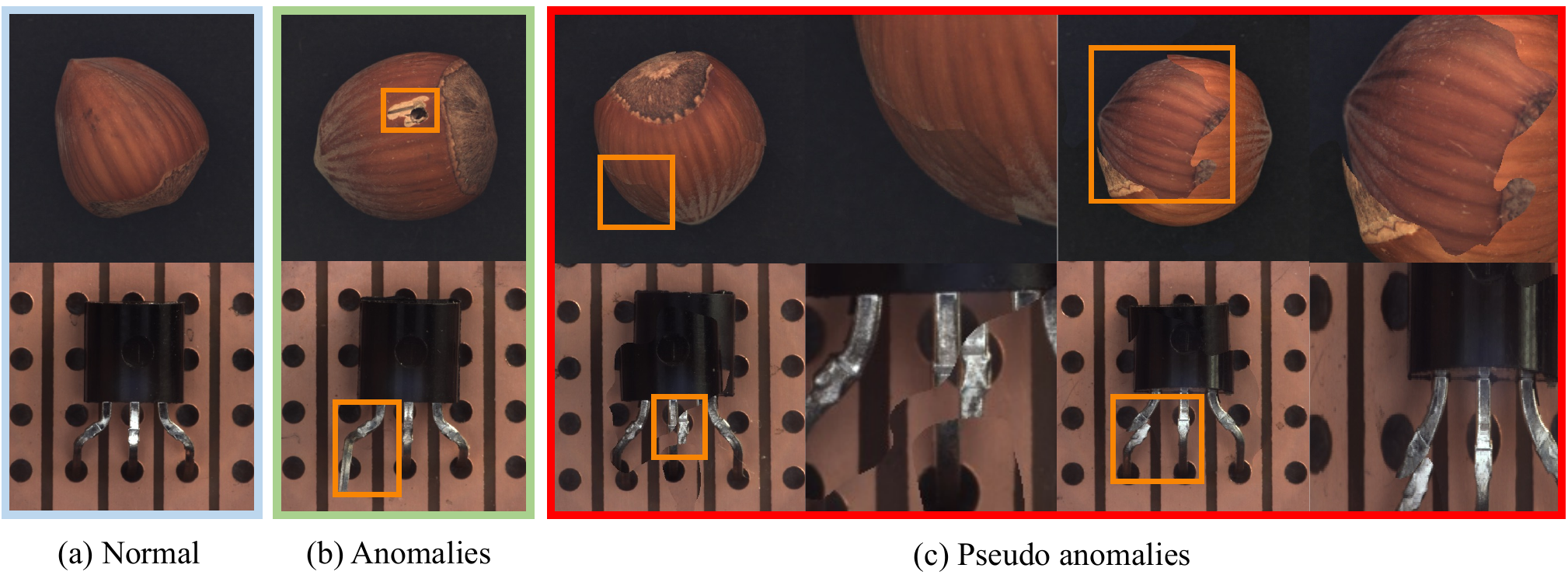}
\caption{Visualization of normal samples, real anomaly samples and pseudo anomaly samples generated by SaliencyCut from Hazelnut and Transistor of MVTec AD dataset~\cite{bergmann2019mvtec}. (a) and (b) show normal and anomaly samples, respectively. Pseudo anomalies generated by SaliencyCut are presented in (c).}
\label{fig:augeddata}
\end{figure*}

To generate such desired pseudo-anomalies, the proposed SaliencyCut is directed by the saliency information.
Since the saliency region of one sample reflects the model's conclusive indication of its type,
the non-saliency region of the normal sample can be regarded as the region where the network believes abnormal information may exist.
To leverage this attribute of non-saliency regions to generate pseudo anomalies, the non-saliency region is pasted into another randomly selected normal sample in the corresponding location.
Figure~\ref{fig:augeddata} provides an example of SaliencyCut.

Algorithm~\ref{alg:algorithm1} presents the augmentation process of SaliencyCut, $\odot$ represents element-wise multiplication in Line~\ref{alg:odot}. To obtain saliency maps, we first input normal training samples and  take the $l_{2}$ norm of gradient values across input channels. The saliency maps are then downsampled to a grid size of $g \times g$ and smoothed to the size of the input image using a B-spline 2D kernel matrix~\cite{lee1982simplified}. Elements of smoothed saliency map distribute continuously  from 0 to 1.  Setting a threshold value in the saliency maps yields a binary mask for isolating non-saliency regions. The threshold is a hyper-parameter that affects the size of non-saliency regions. Threshold selection is described in Section~\ref{exp:sensitive}.

\subsection{Two-Head Learning Mechanism}
Since the generated pseudo-anomalies are well-expand but stay not too further from the distribution of seen anomalies, inspired by ~\cite{ding2022catching}, we design a succinct two-head learning mechanism. Compared with the four-head learning mechanism~\cite{ding2022catching}, our learning mechanism is more concise yet more effective, as later verified both theoretically and empirically.
Generally, normal knowledge can be seen as global features, hence we leverage one normal learning head consisting of a simple MLP structure to learn normal features. As the anomaly information often seems to be fine-grained features, we propose an anomaly learning head consisting of one patch-wise residual head (in Section~\ref{method:patch}) and one top-$K$ scoring module~\cite{pang2019deep, pang2021explainable} to extract anomaly knowledge in the high dimensional semantic space.
Additionally, we present a theoretical interpretation of our two-head learning mechanism.

\textbf{Theoretical Support.} We aim to prove that our two-head learning mechanism offers a lower bound of data log-likelihood   which is tighter compared to the multi-head learning approach~\cite{ding2022catching}.  Suggesting larger evidence, our two-head approach is thus in theory superior to multi-head  in the sense of maximum likelihood. Given a feature extraction network with parameters $f(\cdot,\theta)$, under the two-head learning paradigm, features of the normal instances ${x}_{i}$, anomaly instances ${x}_{j}$ (including seen and pseudo) can be denoted as:
\begin{align}
{Z}_{i}&=f\left({x}_{i} ; {\theta}\right), i \in [1,n],\\
{Q}_{j}&=f\left({x}_{j} ; {\theta}\right), j \in [n+1, n+m],
\end{align}
where $Z_i$ and $Q_j$ are the extracted features of each normal and anomaly instance, respectively. 

\noindent \textbf{Lemma 1:} \emph{Under the two-head learning paradigm, we aim to learn the scoring distribution of  $p(\phi|Z,Q)$, the lower bound of the scoring distribution can be written as:}
\begin{equation}
    \begin{aligned}
    E_{p(z|\phi)}[\log p(\phi|Z,Q)] \ge E_{p(Z|\phi)}[\log {p(Q|\phi)}\\
    -\log{p(Z|\phi)}].
\end{aligned}
\end{equation}

\noindent \textbf{Lemma 2:} \emph{If the anomalies $Q$ are divided into seen anomalies $Q_s$ and pseudo anomalies $A$, and the scoring distribution is learned in a multi-head manner, the lower bound can be written as:}
\begin{equation}
\begin{aligned}
    E_{p(z|\phi)}[\log p(\phi|Z,Q_s,A)] \ge E_{p(Z|\phi)}[\log{p(A|\phi)}\\
    +\log {p(Q_s|\phi)} -\log{p(Z|\phi)}].
\end{aligned}    
% \begin{aligned}
%     E_{p(z|\phi)}[log p(\phi|Z,Q_s,A)] \ge E_{p(Z|\phi)}[log{p(A,Q_s|\phi)}\\
%      -log{p(Z|\phi)}]
% \end{aligned}  
\end{equation}
In the multi-head manner, since the pseudo anomaly instance is derived from normal samples, $A$ can be considered independent of $Q_s$, then we have:
\begin{equation}
\begin{aligned}
   &\log{p(A|\phi)}+\log {p(Q_s|\phi)} \\ 
   &= \log {p(Q_s,A|\phi)}\leq \log{p(Q|\phi)}.
\end{aligned}    
\end{equation}

\begin{figure} 
\centering
\includegraphics[scale=0.4]{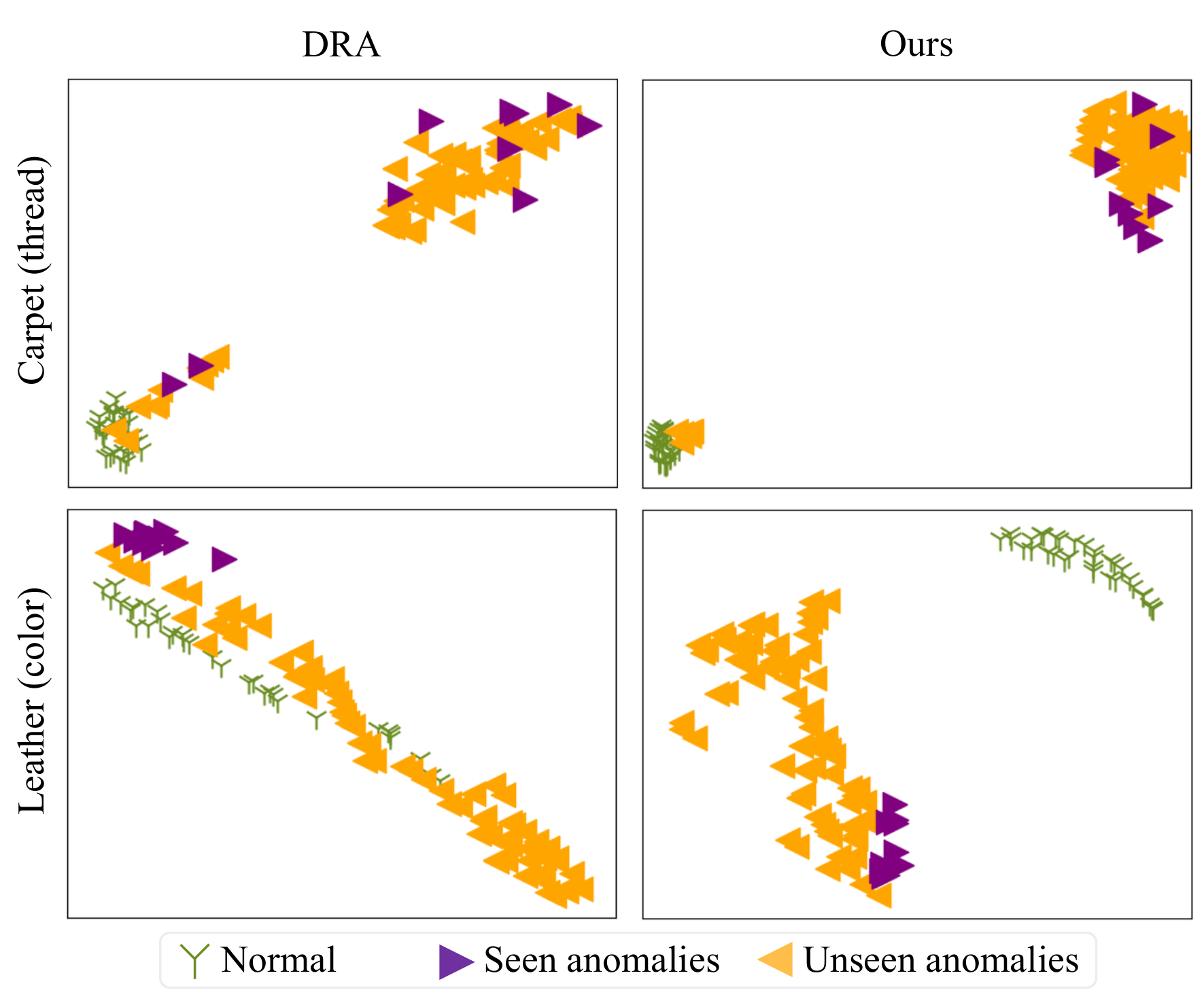}
\caption{t-SNE visualization of features learned by DRA~\cite{ding2022catching} and our approach during testing on two data subsets of MVTec AD~\cite{bergmann2019mvtec} (Leather and Carpet) under the hard setting. Thread and color are two seen anomaly types during training, respectively.}
\label{fig:dtsne}
\end{figure}

Accordingly, our two-head learning mechanism has a tighter lower bound and is easier for optimization than the multi-head approach DRA~\cite{ding2022catching}, which offers theoretical justification for our method. Besides, our approach exhibits superior discriminative ability compared to DRA, as clearly shown in Figure~\ref{fig:dtsne}. Later experiments confirm the advantages of our approach over multi-head approaches.

\subsection{Patch-Wise Residual Module}
\label{method:patch}
We propose a patch-wise residual module for reliable learning of fine-grained anomalous knowledge by making residuals within the samples,
which is motivated by two observations: 1) anomaly samples naturally contain both normal and abnormal features; 2) the residual features within one sample are more semantically uniform compared to that between anomalous and normal samples.

The patch-wise residual module consists of patch splitting and patch-wise subtraction.
After obtaining feature maps from the backbone, we randomly split each feature map into several patches with the same size for residual learning, as shown in the Patch Residual Learning block of Figure~\ref{fig:network}.
Since the location and size of the anomaly area are unknown, the size of each patch is supposed to be large enough to allow anomaly regions to be partitioned in the spitted patches.
Empirically, we set the patches' size  to half the size of the feature map, and one feature map is split into four different patches.
Impact of patch number is described in Section~\ref{exp:sensitive}.
Patch-wise subtraction is performed  within four split patches for residual learning as follows:
\begin{align}
    R = w_{4} - (w_{3} - (w_{2} - w_{1})),
\end{align}
where $R$ denotes residual result and $w_{1}$, $w_{2}$, $w_{3}$ and $w_{4}$ are split patches.
Patch-wise subtraction allows anomalous information to be highlighted in residual results, enabling the network to learn internal anomaly features within samples.

\begin{table*}[ht]
\centering
\setlength{\abovecaptionskip}{3pt}
\caption{AUC results (mean±std) on six real-world datasets under the general setting, with the best results \textbf{highlighted} and the second-best results {\ul underlined}. The first fifteen datasets are data subsets of MVTec AD~\cite{bergmann2019mvtec} whose results are averaged over these subsets.}
\resizebox{\linewidth}{!}{
\begin{tabular}{@{}c|ccccc|ccccc@{}}
\toprule
{\color[HTML]{333333} }                                   & \multicolumn{5}{c|}{{\color[HTML]{333333} \textbf{One Training Anomaly Sample}}}                                                                                                                                                 & \multicolumn{5}{c}{{\color[HTML]{333333} \textbf{Ten Training Anomaly Samples}}}                                                                                                                                                        \\ \cmidrule(l){2-11} 
\multirow{-2}{*}{{\color[HTML]{333333} \textbf{Dataset}}} & {\color[HTML]{333333} \textbf{DevNet}}      & {\color[HTML]{333333} \textbf{MLEP}} & {\color[HTML]{333333} \textbf{DRA}}         & {\color[HTML]{333333} \textbf{SAOE (ensemble)}} & {\color[HTML]{333333} \textbf{Ours}}        & {\color[HTML]{333333} \textbf{DevNet}}      & {\color[HTML]{333333} \textbf{MLEP}}        & {\color[HTML]{333333} \textbf{DRA}}         & {\color[HTML]{333333} \textbf{SAOE (ensemble)}} & {\color[HTML]{333333} \textbf{Ours}}        \\ \midrule
{\color[HTML]{333333} Carpet}                             & {\color[HTML]{333333} 0.746±0.076}          & {\color[HTML]{333333} 0.701±0.091}   & {\color[HTML]{333333} {\ul 0.859±0.023}}    & {\color[HTML]{333333} 0.766±0.098}              & {\color[HTML]{333333} \textbf{0.957±0.014}} & {\color[HTML]{333333} 0.867±0.040}          & {\color[HTML]{333333} 0.781±0.049}          & {\color[HTML]{333333} {\ul 0.940±0.027}}    & {\color[HTML]{333333} 0.755±0.136}              & {\color[HTML]{333333} \textbf{0.971±0.007}} \\
{\color[HTML]{333333} Grid}                               & {\color[HTML]{333333} 0.891±0.040}          & {\color[HTML]{333333} 0.839±0.028}   & {\color[HTML]{333333} {\ul 0.972±0.011}}    & {\color[HTML]{333333} 0.921±0.032}              & {\color[HTML]{333333} \textbf{0.990±0.004}} & {\color[HTML]{333333} 0.967±0.021}          & {\color[HTML]{333333} 0.980±0.009}          & {\color[HTML]{333333} {\ul 0.987±0.009}}    & {\color[HTML]{333333} 0.952±0.011}              & {\color[HTML]{333333} \textbf{1.000±0.000}} \\
{\color[HTML]{333333} Leather}                            & {\color[HTML]{333333} 0.873±0.026}          & {\color[HTML]{333333} 0.781±0.020}   & {\color[HTML]{333333} 0.989±0.005}          & {\color[HTML]{333333} {\ul 0.996±0.007}}        & {\color[HTML]{333333} \textbf{0.997±0.002}} & {\color[HTML]{333333} {\ul 0.999±0.001}}    & {\color[HTML]{333333} 0.813±0.158}          & {\color[HTML]{333333} \textbf{1.000±0.000}} & {\color[HTML]{333333} \textbf{1.000±0.000}}     & {\color[HTML]{333333} \textbf{1.000±0.000}} \\
{\color[HTML]{333333} Tile}                               & {\color[HTML]{333333} 0.752±0.038}          & {\color[HTML]{333333} 0.927±0.036}   & {\color[HTML]{333333} {\ul 0.965±0.015}}    & {\color[HTML]{333333} 0.935±0.034}              & {\color[HTML]{333333} \textbf{0.988±0.008}} & {\color[HTML]{333333} 0.987±0.005}          & {\color[HTML]{333333} 0.988±0.009}          & {\color[HTML]{333333} {\ul 0.994±0.006}}    & {\color[HTML]{333333} 0.944±0.013}              & {\color[HTML]{333333} \textbf{1.000±0.000}} \\
{\color[HTML]{333333} Wood}                               & {\color[HTML]{333333} 0.900±0.068}          & {\color[HTML]{333333} 0.660±0.142}   & {\color[HTML]{333333} {\ul 0.985±0.011}}    & {\color[HTML]{333333} 0.948±0.009}              & {\color[HTML]{333333} \textbf{0.993±0.003}} & {\color[HTML]{333333} \textbf{0.999±0.001}} & {\color[HTML]{333333} \textbf{0.999±0.002}} & {\color[HTML]{333333} {\ul 0.998±0.001}}    & {\color[HTML]{333333} 0.976±0.031}              & {\color[HTML]{333333} {\ul 0.998±0.001}}    \\
{\color[HTML]{333333} Bottle}                             & {\color[HTML]{333333} 0.976±0.006}          & {\color[HTML]{333333} 0.927±0.090}   & {\color[HTML]{333333} \textbf{1.000±0.000}} & {\color[HTML]{333333} {\ul 0.989±0.019}}        & {\color[HTML]{333333} \textbf{1.000±0.000}} & {\color[HTML]{333333} 0.993±0.008}          & {\color[HTML]{333333} 0.981±0.004}          & {\color[HTML]{333333} \textbf{1.000±0.000}} & {\color[HTML]{333333} {\ul 0.998±0.003}}        & {\color[HTML]{333333} \textbf{1.000±0.000}} \\
{\color[HTML]{333333} Capsule}                            & {\color[HTML]{333333} 0.564±0.032}          & {\color[HTML]{333333} 0.558±0.075}   & {\color[HTML]{333333} {\ul 0.631±0.056}}    & {\color[HTML]{333333} 0.611±0.109}              & {\color[HTML]{333333} \textbf{0.850±0.019}} & {\color[HTML]{333333} 0.865±0.057}          & {\color[HTML]{333333} 0.818±0.063}          & {\color[HTML]{333333} {\ul 0.935±0.022}}    & {\color[HTML]{333333} 0.850±0.054}              & {\color[HTML]{333333} \textbf{0.969±0.002}} \\
{\color[HTML]{333333} Pill}                               & {\color[HTML]{333333} 0.769±0.017}          & {\color[HTML]{333333} 0.656±0.061}   & {\color[HTML]{333333} {\ul 0.832±0.034}}    & {\color[HTML]{333333} 0.652±0.078}              & {\color[HTML]{333333} \textbf{0.906±0.007}} & {\color[HTML]{333333} 0.866±0.038}          & {\color[HTML]{333333} 0.845±0.048}          & {\color[HTML]{333333} {\ul 0.904±0.024}}    & {\color[HTML]{333333} 0.872±0.049}              & {\color[HTML]{333333} \textbf{0.944±0.010}} \\
{\color[HTML]{333333} Transistor}                         & {\color[HTML]{333333} {\ul 0.722±0.032}}    & {\color[HTML]{333333} 0.695±0.124}   & {\color[HTML]{333333} 0.668±0.068}          & {\color[HTML]{333333} 0.680±0.182}              & {\color[HTML]{333333} \textbf{0.754±0.015}} & {\color[HTML]{333333} {\ul 0.924±0.027}}    & {\color[HTML]{333333} \textbf{0.927±0.043}} & {\color[HTML]{333333} 0.915±0.025}          & {\color[HTML]{333333} 0.860±0.053}              & {\color[HTML]{333333} 0.917±0.029}          \\
{\color[HTML]{333333} Zipper}                             & {\color[HTML]{333333} 0.922±0.018}          & {\color[HTML]{333333} 0.856±0.086}   & {\color[HTML]{333333} {\ul 0.984±0.016}}    & {\color[HTML]{333333} 0.970±0.033}              & {\color[HTML]{333333} \textbf{0.998±0.001}} & {\color[HTML]{333333} 0.990±0.009}          & {\color[HTML]{333333} 0.965±0.002}          & {\color[HTML]{333333} \textbf{1.000±0.000}} & {\color[HTML]{333333} {\ul 0.995±0.004}}        & {\color[HTML]{333333} \textbf{1.000±0.000}} \\
{\color[HTML]{333333} Cable}                              & {\color[HTML]{333333} 0.783±0.058}          & {\color[HTML]{333333} 0.688±0.017}   & {\color[HTML]{333333} \textbf{0.876±0.012}} & {\color[HTML]{333333} {\ul 0.819±0.060}}        & {\color[HTML]{333333} \textbf{0.876±0.013}} & {\color[HTML]{333333} 0.892±0.020}          & {\color[HTML]{333333} 0.857±0.062}          & {\color[HTML]{333333} {\ul 0.909±0.011}}    & {\color[HTML]{333333} 0.862±0.022}              & {\color[HTML]{333333} \textbf{0.940±0.006}} \\
{\color[HTML]{333333} Hazelnut}                           & {\color[HTML]{333333} {\ul 0.979±0.010}}    & {\color[HTML]{333333} 0.704±0.090}   & {\color[HTML]{333333} 0.977±0.030}          & {\color[HTML]{333333} 0.961±0.042}              & {\color[HTML]{333333} \textbf{0.987±0.005}} & {\color[HTML]{333333} \textbf{1.000±0.000}} & {\color[HTML]{333333} \textbf{1.000±0.000}} & {\color[HTML]{333333} \textbf{1.000±0.000}} & {\color[HTML]{333333} \textbf{1.000±0.000}}     & {\color[HTML]{333333} \textbf{1.000±0.000}} \\
{\color[HTML]{333333} Metal\_nut}                         & {\color[HTML]{333333} 0.876±0.007}          & {\color[HTML]{333333} 0.878±0.038}   & {\color[HTML]{333333} {\ul 0.948±0.046}}    & {\color[HTML]{333333} 0.922±0.033}              & {\color[HTML]{333333} \textbf{0.966±0.000}} & {\color[HTML]{333333} 0.991±0.006}          & {\color[HTML]{333333} 0.974±0.009}          & {\color[HTML]{333333} {\ul 0.997±0.002}}    & {\color[HTML]{333333} 0.976±0.013}              & {\color[HTML]{333333} \textbf{0.999±0.000}} \\
{\color[HTML]{333333} Screw}                              & {\color[HTML]{333333} 0.399±0.187}          & {\color[HTML]{333333} 0.675±0.294}   & {\color[HTML]{333333} {\ul 0.903±0.064}}    & {\color[HTML]{333333} 0.653±0.074}              & {\color[HTML]{333333} \textbf{0.961±0.011}} & {\color[HTML]{333333} 0.970±0.015}          & {\color[HTML]{333333} 0.899±0.039}          & {\color[HTML]{333333} {\ul 0.977±0.009}}    & {\color[HTML]{333333} 0.975±0.023}              & {\color[HTML]{333333} \textbf{0.990±0.005}} \\
{\color[HTML]{333333} Toothbrush}                         & {\color[HTML]{333333} \textbf{0.753±0.027}} & {\color[HTML]{333333} 0.617±0.058}   & {\color[HTML]{333333} 0.650±0.029}          & {\color[HTML]{333333} 0.686±0.110}              & {\color[HTML]{333333} {\ul 0.752±0.009}}    & {\color[HTML]{333333} 0.860±0.066}          & {\color[HTML]{333333} 0.783±0.048}          & {\color[HTML]{333333} 0.826±0.021}          & {\color[HTML]{333333} {\ul 0.865±0.062}}        & {\color[HTML]{333333} \textbf{0.909±0.005}} \\ \midrule
{\color[HTML]{333333} \textbf{MVTec AD}}                  & {\color[HTML]{333333} 0.794±0.014}          & {\color[HTML]{333333} 0.744±0.019}   & {\color[HTML]{333333} {\ul 0.883±0.008}}    & {\color[HTML]{333333} 0.834±0.007}              & {\color[HTML]{333333} \textbf{0.932±0.083}} & {\color[HTML]{333333} 0.945±0.004}          & {\color[HTML]{333333} 0.907±0.005}          & {\color[HTML]{333333} {\ul 0.959±0.003}}    & {\color[HTML]{333333} 0.926±0.010}              & {\color[HTML]{333333} \textbf{0.976±0.032}} \\
{\color[HTML]{333333} \textbf{Mastcam}}                   & {\color[HTML]{333333} 0.595±0.016}          & {\color[HTML]{333333} 0.625±0.045}   & {\color[HTML]{333333} {\ul 0.692±0.058}}    & {\color[HTML]{333333} 0.662±0.018}              & {\color[HTML]{333333} \textbf{0.694±0.011}} & {\color[HTML]{333333} 0.790±0.021}          & {\color[HTML]{333333} 0.798±0.026}          & {\color[HTML]{333333} \textbf{0.848±0.008}} & {\color[HTML]{333333} 0.810±0.029}              & {\color[HTML]{333333} {\ul 0.827±0.035}}    \\
{\color[HTML]{333333} \textbf{SDD}}                       & {\color[HTML]{333333} {\ul 0.881±0.009}}    & {\color[HTML]{333333} 0.811±0.045}   & {\color[HTML]{333333} 0.859±0.014}          & {\color[HTML]{333333} 0.781±0.009}              & {\color[HTML]{333333} \textbf{0.897±0.003}} & {\color[HTML]{333333} {\ul 0.988±0.006}}    & {\color[HTML]{333333} 0.983±0.013}          & {\color[HTML]{333333} \textbf{0.991±0.005}} & {\color[HTML]{333333} 0.955±0.020}              & {\color[HTML]{333333} \textbf{0.991±0.001}} \\
{\color[HTML]{333333} \textbf{ELPV}}                      & {\color[HTML]{333333} 0.514±0.076}          & {\color[HTML]{333333} 0.578±0.062}   & {\color[HTML]{333333} {\ul 0.675±0.024}}    & {\color[HTML]{333333} 0.635±0.092}              & {\color[HTML]{333333} \textbf{0.699±0.051}} & {\color[HTML]{333333} {\ul 0.846±0.022}}    & {\color[HTML]{333333} 0.794±0.047}          & {\color[HTML]{333333} 0.845±0.013}          & {\color[HTML]{333333} 0.793±0.047}              & {\color[HTML]{333333} \textbf{0.852±0.005}} \\
{\color[HTML]{333333} \textbf{Optical}}                   & {\color[HTML]{333333} 0.523±0.003}          & {\color[HTML]{333333} 0.516±0.009}   & {\color[HTML]{333333} {\ul 0.888±0.012}}    & {\color[HTML]{333333} 0.815±0.014}              & {\color[HTML]{333333} \textbf{0.895±0.009}} & {\color[HTML]{333333} 0.782±0.065}          & {\color[HTML]{333333} 0.740±0.039}          & {\color[HTML]{333333} \textbf{0.965±0.006}} & {\color[HTML]{333333} {\ul 0.941±0.013}}        & {\color[HTML]{333333} \textbf{0.965±0.001}} \\
{\color[HTML]{333333} \textbf{AITEX}}                     & {\color[HTML]{333333} 0.598±0.070}          & {\color[HTML]{333333} 0.564±0.055}   & {\color[HTML]{333333} {\ul 0.692±0.124}}    & {\color[HTML]{333333} 0.675±0.094}              & {\color[HTML]{333333} \textbf{0.719±0.035}} & {\color[HTML]{333333} 0.887±0.013}          & {\color[HTML]{333333} 0.867±0.037}          & {\color[HTML]{333333} {\ul 0.893±0.017}}    & {\color[HTML]{333333} 0.874±0.024}              & {\color[HTML]{333333} \textbf{0.911±0.008}} \\ \bottomrule
\end{tabular}}
\label{tab:general}
\end{table*}

%------------------------------------------------------------------------
\section{Experiments}
\subsection{Datasets}
In experiments, we evaluate our method on six datasets with real anomalies:
five industrial defect detection datasets include MVTec AD~\cite{bergmann2019mvtec},  SDD~\cite{tabernik2020segmentation}, AITEX~\cite{silvestre2019public}, ELPV~\cite{deitsch2019automatic}, and Optical~\cite{wieler2007weakly}, while one dataset Mastcam describes geological phenomena on Mars which regards particular geological features as anomalies~\cite{kerner2020comparison}.

\subsection{Implementation Details}
ResNet-18~\cite{he2016deep} serves as the backbone network for feature learning.  During training, the model is trained for 30 epochs, with 20 iterations per epoch and a batch size of 48, in which parameters are optimized by Adam.
We empirically set the grid size $g$ to 18 and the threshold for obtaining saliency masks to 0.4.
The top-$K$ anomaly scoring approach is identical to that of DevNet~\cite{pang2019deep, pang2021explainable} and DRA~\cite{ding2022catching} with $K$ set to 10\%.
Area Under Receiver Operating Characteristic Curve (AUC-ROC) is utilized as the performance evaluation metric.
All reported results are averaged over three independent trials.

Our model is compared to the four most related state-of-the-art supervised anomaly detection methods that use limited anomalies during training, including MLEP~\cite{liu2019margin}, DevNet~\cite{pang2019deep, pang2021explainable}, DRA~\cite{ding2022catching} and one ensemble method SAOE~\cite{ding2022catching} (combining data augmentation-based \underline{S}ynthetic \underline{A}nomalies~\cite{li2021cutpaste,liznerski2020explainable,tack2020csi} with \underline{O}utlier \underline{E}xposure~\cite{hendrycks2018deep,reiss2021panda}).
DevNet, MLEP and DRA also attempt to tackle the open-set fine-grained anomaly detection issue. SAOE performs data augmentation for pseudo anomalies generation and outlier exposure methods for anomaly detection, which is superior to utilizing one of these pseudo anomalies generation methods.
The results of DevNet, MLEP, DRA and SAOE are obtained from DRA~\cite{ding2022catching}.
The results of three anomaly-free detection methods, KDAD~\cite{salehi2021multiresolution}, SSM~\cite{huang2022self} and CutPaste~\cite{li2021cutpaste}, are also from their papers.

\subsection{Experimental Settings}
{\bf General Setting.} In the general setting of open-set fine-grained anomaly detection, a few anomalies randomly selected from the test set are accessible during training, and the test set contains both seen anomalies and unseen anomalies or only seen anomalies.
To measure the performance of the model with a few anomaly samples and very few anomaly samples, respectively, we set the general setting experiment with one training anomaly sample and ten training anomaly samples.

{\bf Hard Setting.} In the hard setting, only one type of anomaly drawn from the test set is accessible during training, while the test set only contains unseen types of anomalies.
Hard setting experiment is also set with one and ten anomaly training samples, respectively.

{\bf Anomaly-Free Setting.} In the anomaly-free setting, there are no anomaly samples during training, the test set contains all types of anomalies.

\begin{table*}[ht]
\centering
\setlength{\abovecaptionskip}{3pt}

\caption{AUC results under the hard setting, where models are trained with one known anomaly type and tested to detect the rest of all other anomaly types. Each data subset is named by the known anomaly type.}
\resizebox{\linewidth}{!}{
\begin{tabular}{@{}cc|ccccc|ccccc@{}}
\toprule
{\color[HTML]{333333} }                                      & {\color[HTML]{333333} }                                       & \multicolumn{5}{c|}{{\color[HTML]{333333} \textbf{One Training Anomaly Sample}}}                                                                                                                                                 & \multicolumn{5}{c}{{\color[HTML]{333333} \textbf{Ten Training Anomlay Samples}}}                                                                                                                                                        \\ \cmidrule(l){3-12} 
\multirow{-2}{*}{{\color[HTML]{333333} \textbf{Dataset}}}    & \multirow{-2}{*}{{\color[HTML]{333333} \textbf{Data Subset}}} & {\color[HTML]{333333} \textbf{DevNet}}      & {\color[HTML]{333333} \textbf{MLEP}} & {\color[HTML]{333333} \textbf{DRA}}         & {\color[HTML]{333333} \textbf{SAOE (ensemble)}} & {\color[HTML]{333333} \textbf{Ours}}        & {\color[HTML]{333333} \textbf{DevNet}}      & {\color[HTML]{333333} \textbf{MLEP}}        & {\color[HTML]{333333} \textbf{DRA}}         & {\color[HTML]{333333} \textbf{SAOE (ensemble)}} & {\color[HTML]{333333} \textbf{Ours}}        \\ \midrule
{\color[HTML]{333333} }                                      & {\color[HTML]{333333} Color}                                  & {\color[HTML]{333333} 0.716±0.085}          & {\color[HTML]{333333} 0.547±0.056}   & {\color[HTML]{333333} {\ul 0.879±0.021}}    & {\color[HTML]{333333} 0.763±0.100}              & {\color[HTML]{333333} \textbf{0.932±0.019}} & {\color[HTML]{333333} 0.767±0.015}          & {\color[HTML]{333333} 0.698±0.025}          & {\color[HTML]{333333} {\ul 0.886±0.042}}    & {\color[HTML]{333333} 0.467±0.067}              & {\color[HTML]{333333} \textbf{0.951±0.005}} \\
{\color[HTML]{333333} }                                      & {\color[HTML]{333333} Cut}                                    & {\color[HTML]{333333} 0.666±0.035}          & {\color[HTML]{333333} 0.658±0.056}   & {\color[HTML]{333333} {\ul 0.902±0.033}}    & {\color[HTML]{333333} 0.664±0.165}              & {\color[HTML]{333333} \textbf{0.968±0.005}} & {\color[HTML]{333333} 0.819±0.037}          & {\color[HTML]{333333} 0.653±0.120}          & {\color[HTML]{333333} {\ul 0.922±0.038}}    & {\color[HTML]{333333} 0.793±0.175}              & {\color[HTML]{333333} \textbf{0.964±0.008}} \\
{\color[HTML]{333333} }                                      & {\color[HTML]{333333} Hole}                                   & {\color[HTML]{333333} 0.721±0.067}          & {\color[HTML]{333333} 0.653±0.065}   & {\color[HTML]{333333} {\ul 0.901±0.033}}    & {\color[HTML]{333333} 0.772±0.071}              & {\color[HTML]{333333} \textbf{0.967±0.005}} & {\color[HTML]{333333} 0.814±0.038}          & {\color[HTML]{333333} 0.674±0.076}          & {\color[HTML]{333333} {\ul 0.947±0.016}}    & {\color[HTML]{333333} 0.831±0.125}              & {\color[HTML]{333333} \textbf{0.961±0.004}} \\
{\color[HTML]{333333} }                                      & {\color[HTML]{333333} Metal}                                  & {\color[HTML]{333333} 0.819±0.032}          & {\color[HTML]{333333} 0.706±0.047}   & {\color[HTML]{333333} {\ul 0.871±0.037}}    & {\color[HTML]{333333} 0.780±0.172}              & {\color[HTML]{333333} \textbf{0.939±0.009}} & {\color[HTML]{333333} 0.863±0.022}          & {\color[HTML]{333333} 0.764±0.061}          & {\color[HTML]{333333} {\ul 0.933±0.022}}    & {\color[HTML]{333333} 0.883±0.043}              & {\color[HTML]{333333} \textbf{0.951±0.002}} \\
{\color[HTML]{333333} }                                      & {\color[HTML]{333333} Thread}                                 & {\color[HTML]{333333} 0.912±0.044}          & {\color[HTML]{333333} 0.831±0.117}   & {\color[HTML]{333333} {\ul 0.950±0.029}}    & {\color[HTML]{333333} 0.787±0.204}              & {\color[HTML]{333333} \textbf{0.994±0.001}} & {\color[HTML]{333333} 0.972±0.009}          & {\color[HTML]{333333} 0.967±0.006}          & {\color[HTML]{333333} {\ul 0.989±0.004}}    & {\color[HTML]{333333} 0.834±0.297}              & {\color[HTML]{333333} \textbf{0.998±0.000}} \\ \cmidrule(l){2-12} 
\multirow{-6}{*}{{\color[HTML]{333333} \textbf{Carpet}}}     & {\color[HTML]{333333} \textbf{Mean}}                          & {\color[HTML]{333333} 0.767±0.018}          & {\color[HTML]{333333} 0.679±0.029}   & {\color[HTML]{333333} {\ul 0.901±0.006}}    & {\color[HTML]{333333} 0.753±0.055}              & {\color[HTML]{333333} \textbf{0.960±0.024}} & {\color[HTML]{333333} 0.847±0.017}          & {\color[HTML]{333333} 0.751±0.023}          & {\color[HTML]{333333} {\ul 0.935±0.013}}    & {\color[HTML]{333333} 0.762±0.073}              & {\color[HTML]{333333} \textbf{0.965±0.018}} \\ \midrule
{\color[HTML]{333333} }                                      & {\color[HTML]{333333} Bent}                                   & {\color[HTML]{333333} 0.797±0.048}          & {\color[HTML]{333333} 0.743±0.013}   & {\color[HTML]{333333} {\ul 0.952±0.020}}    & {\color[HTML]{333333} 0.864±0.032}              & {\color[HTML]{333333} \textbf{0.976±0.012}} & {\color[HTML]{333333} 0.904±0.022}          & {\color[HTML]{333333} 0.956±0.013}          & {\color[HTML]{333333} {\ul 0.990±0.003}}    & {\color[HTML]{333333} 0.901±0.023}              & {\color[HTML]{333333} \textbf{0.996±0.003}} \\
{\color[HTML]{333333} }                                      & {\color[HTML]{333333} Color}                                  & {\color[HTML]{333333} 0.909±0.023}          & {\color[HTML]{333333} 0.835±0.075}   & {\color[HTML]{333333} {\ul 0.946±0.023}}    & {\color[HTML]{333333} 0.857±0.037}              & {\color[HTML]{333333} \textbf{0.967±0.019}} & {\color[HTML]{333333} {\ul 0.978±0.016}}    & {\color[HTML]{333333} 0.945±0.039}          & {\color[HTML]{333333} 0.967±0.011}          & {\color[HTML]{333333} 0.879±0.018}              & {\color[HTML]{333333} \textbf{0.979±0.004}} \\
{\color[HTML]{333333} }                                      & {\color[HTML]{333333} Flip}                                   & {\color[HTML]{333333} 0.764±0.014}          & {\color[HTML]{333333} 0.813±0.031}   & {\color[HTML]{333333} {\ul 0.921±0.029}}    & {\color[HTML]{333333} 0.751±0.090}              & {\color[HTML]{333333} \textbf{0.953±0.014}} & {\color[HTML]{333333} \textbf{0.987±0.004}} & {\color[HTML]{333333} 0.805±0.057}          & {\color[HTML]{333333} 0.913±0.021}          & {\color[HTML]{333333} 0.795±0.062}              & {\color[HTML]{333333} {\ul 0.950±0.013}}    \\
{\color[HTML]{333333} }                                      & {\color[HTML]{333333} Scratch}                                & {\color[HTML]{333333} \textbf{0.952±0.052}} & {\color[HTML]{333333} 0.907±0.085}   & {\color[HTML]{333333} 0.909±0.023}          & {\color[HTML]{333333} 0.792±0.075}              & {\color[HTML]{333333} {\ul 0.948±0.006}}    & {\color[HTML]{333333} \textbf{0.991±0.017}} & {\color[HTML]{333333} 0.805±0.153}          & {\color[HTML]{333333} 0.911±0.034}          & {\color[HTML]{333333} 0.845±0.041}              & {\color[HTML]{333333} {\ul 0.983±0.002}}    \\ \cmidrule(l){2-12} 
\multirow{-5}{*}{{\color[HTML]{333333} \textbf{Metal\_nut}}} & {\color[HTML]{333333} \textbf{Mean}}                          & {\color[HTML]{333333} 0.855±0.016}          & {\color[HTML]{333333} 0.825±0.023}   & {\color[HTML]{333333} {\ul 0.932±0.017}}    & {\color[HTML]{333333} 0.816±0.029}              & {\color[HTML]{333333} \textbf{0.961±0.017}} & {\color[HTML]{333333} {\ul 0.965±0.011}}    & {\color[HTML]{333333} 0.878±0.058}          & {\color[HTML]{333333} 0.945±0.017}          & {\color[HTML]{333333} 0.855±0.016}              & {\color[HTML]{333333} \textbf{0.977±0.018}} \\ \midrule
                                                             & {\color[HTML]{333333} Bedrock}                                & {\color[HTML]{333333} 0.495±0.028}          & {\color[HTML]{333333} 0.532±0.036}   & {\color[HTML]{333333} \textbf{0.668±0.012}} & {\color[HTML]{333333} {\ul 0.636±0.072}}        & {\color[HTML]{333333} 0.626±0.037}          & {\color[HTML]{333333} 0.550±0.053}          & {\color[HTML]{333333} 0.512±0.062}          & {\color[HTML]{333333} \textbf{0.658±0.021}} & {\color[HTML]{333333} {\ul 0.636±0.068}}        & {\color[HTML]{333333} 0.581±0.054}          \\
                                                             & {\color[HTML]{333333} Broken-rock}                            & {\color[HTML]{333333} 0.533±0.020}          & {\color[HTML]{333333} 0.544±0.088}   & {\color[HTML]{333333} 0.645±0.053}          & {\color[HTML]{333333} \textbf{0.699±0.058}}     & {\color[HTML]{333333} {\ul 0.674±0.058}}    & {\color[HTML]{333333} 0.547±0.018}          & {\color[HTML]{333333} 0.651±0.063}          & {\color[HTML]{333333} 0.649±0.047}          & {\color[HTML]{333333} \textbf{0.712±0.052}}     & {\color[HTML]{333333} {\ul 0.684±0.013}}    \\
                                                             & {\color[HTML]{333333} Drill-hole}                             & {\color[HTML]{333333} 0.555±0.037}          & {\color[HTML]{333333} 0.636±0.066}   & {\color[HTML]{333333} 0.657±0.070}          & {\color[HTML]{333333} \textbf{0.697±0.074}}     & {\color[HTML]{333333} {\ul 0.665±0.022}}    & {\color[HTML]{333333} 0.583±0.022}          & {\color[HTML]{333333} 0.660±0.002}          & {\color[HTML]{333333} {\ul 0.725±0.005}}    & {\color[HTML]{333333} 0.682±0.042}              & {\color[HTML]{333333} \textbf{0.773±0.017}} \\
                                                             & {\color[HTML]{333333} Drt}                                    & {\color[HTML]{333333} 0.529±0.046}          & {\color[HTML]{333333} 0.624±0.042}   & {\color[HTML]{333333} {\ul 0.713±0.053}}    & {\color[HTML]{333333} \textbf{0.735±0.020}}     & {\color[HTML]{333333} 0.698±0.044}          & {\color[HTML]{333333} 0.621±0.043}          & {\color[HTML]{333333} 0.616±0.048}          & {\color[HTML]{333333} 0.760±0.033}          & {\color[HTML]{333333} {\ul 0.761±0.062}}        & {\color[HTML]{333333} \textbf{0.796±0.028}} \\
                                                             & {\color[HTML]{333333} Dump-pile}                              & {\color[HTML]{333333} 0.521±0.020}          & {\color[HTML]{333333} 0.545±0.127}   & {\color[HTML]{333333} {\ul 0.767±0.043}}    & {\color[HTML]{333333} 0.682±0.022}              & {\color[HTML]{333333} \textbf{0.775±0.020}} & {\color[HTML]{333333} 0.705±0.011}          & {\color[HTML]{333333} 0.696±0.047}          & {\color[HTML]{333333} 0.748±0.066}          & {\color[HTML]{333333} {\ul 0.750±0.037}}        & {\color[HTML]{333333} \textbf{0.773±0.053}} \\
                                                             & {\color[HTML]{333333} Float}                                  & {\color[HTML]{333333} 0.502±0.020}          & {\color[HTML]{333333} 0.530±0.075}   & {\color[HTML]{333333} {\ul 0.670±0.065}}    & {\color[HTML]{333333} \textbf{0.711±0.041}}     & {\color[HTML]{333333} 0.631±0.048}          & {\color[HTML]{333333} 0.615±0.052}          & {\color[HTML]{333333} 0.671±0.032}          & {\color[HTML]{333333} \textbf{0.744±0.073}} & {\color[HTML]{333333} 0.718±0.064}              & {\color[HTML]{333333} {\ul 0.735±0.007}}    \\
                                                             & {\color[HTML]{333333} Meteorite}                              & {\color[HTML]{333333} 0.467±0.049}          & {\color[HTML]{333333} 0.476±0.014}   & {\color[HTML]{333333} 0.637±0.015}          & {\color[HTML]{333333} {\ul 0.669±0.037}}        & {\color[HTML]{333333} \textbf{0.712±0.026}} & {\color[HTML]{333333} 0.554±0.021}          & {\color[HTML]{333333} 0.473±0.047}          & {\color[HTML]{333333} {\ul 0.716±0.004}}    & {\color[HTML]{333333} 0.647±0.030}              & {\color[HTML]{333333} \textbf{0.721±0.066}} \\
                                                             & {\color[HTML]{333333} Scuff}                                  & {\color[HTML]{333333} 0.472±0.031}          & {\color[HTML]{333333} 0.492±0.037}   & {\color[HTML]{333333} 0.549±0.027}          & {\color[HTML]{333333} \textbf{0.679±0.048}}     & {\color[HTML]{333333} {\ul 0.588±0.040}}    & {\color[HTML]{333333} 0.528±0.034}          & {\color[HTML]{333333} 0.504±0.052}          & {\color[HTML]{333333} {\ul 0.636±0.086}}    & {\color[HTML]{333333} \textbf{0.676±0.019}}     & {\color[HTML]{333333} 0.607±0.021}          \\
                                                             & {\color[HTML]{333333} Veins}                                  & {\color[HTML]{333333} 0.527±0.023}          & {\color[HTML]{333333} 0.489±0.028}   & {\color[HTML]{333333} \textbf{0.699±0.045}} & {\color[HTML]{333333} {\ul 0.688±0.069}}        & {\color[HTML]{333333} 0.661±0.043}          & {\color[HTML]{333333} 0.589±0.072}          & {\color[HTML]{333333} 0.510±0.090}          & {\color[HTML]{333333} 0.620±0.036}          & {\color[HTML]{333333} \textbf{0.686±0.053}}     & {\color[HTML]{333333} {\ul 0.660±0.007}}    \\ \cmidrule(l){2-12} 
\multirow{-10}{*}{\textbf{MastCam}}                          & {\color[HTML]{333333} \textbf{Mean}}                          & {\color[HTML]{333333} 0.511±0.013}          & {\color[HTML]{333333} 0.541±0.007}   & {\color[HTML]{333333} 0.667±0.012}          & {\color[HTML]{333333} \textbf{0.689±0.037}}     & {\color[HTML]{333333} {\ul 0.670±0.066}}    & {\color[HTML]{333333} 0.588±0.011}          & {\color[HTML]{333333} 0.588±0.016}          & {\color[HTML]{333333} 0.695±0.004}          & {\color[HTML]{333333} {\ul 0.697±0.014}}        & {\color[HTML]{333333} \textbf{0.703±0.080}} \\ \midrule
{\color[HTML]{333333} }                                      & {\color[HTML]{333333} Mono}                                   & {\color[HTML]{333333} 0.634±0.087}          & {\color[HTML]{333333} 0.649±0.027}   & {\color[HTML]{333333} \textbf{0.735±0.031}} & {\color[HTML]{333333} 0.563±0.102}              & {\color[HTML]{333333} {\ul 0.720±0.054}}    & {\color[HTML]{333333} 0.599±0.040}          & {\color[HTML]{333333} \textbf{0.756±0.045}} & {\color[HTML]{333333} 0.731±0.021}          & {\color[HTML]{333333} 0.569±0.035}              & {\color[HTML]{333333} {\ul 0.735±0.020}}    \\
{\color[HTML]{333333} }                                      & {\color[HTML]{333333} Poly}                                   & {\color[HTML]{333333} 0.662±0.050}          & {\color[HTML]{333333} 0.483±0.247}   & {\color[HTML]{333333} {\ul 0.671±0.051}}    & {\color[HTML]{333333} 0.665±0.173}              & {\color[HTML]{333333} \textbf{0.734±0.074}} & {\color[HTML]{333333} {\ul 0.804±0.022}}    & {\color[HTML]{333333} 0.734±0.078}          & {\color[HTML]{333333} 0.800±0.064}          & {\color[HTML]{333333} 0.796±0.084}              & {\color[HTML]{333333} \textbf{0.827±0.052}} \\ \cmidrule(l){2-12} 
\multirow{-3}{*}{{\color[HTML]{333333} \textbf{ELPV}}}       & {\color[HTML]{333333} \textbf{Mean}}                          & {\color[HTML]{333333} 0.648±0.057}          & {\color[HTML]{333333} 0.566±0.111}   & {\color[HTML]{333333} {\ul 0.703±0.022}}    & {\color[HTML]{333333} 0.614±0.048}              & {\color[HTML]{333333} \textbf{0.727±0.065}} & {\color[HTML]{333333} 0.702±0.023}          & {\color[HTML]{333333} 0.745±0.020}          & {\color[HTML]{333333} {\ul 0.766±0.029}}    & {\color[HTML]{333333} 0.683±0.047}              & {\color[HTML]{333333} \textbf{0.781±0.060}} \\ \midrule
{\color[HTML]{333333} }                                      & {\color[HTML]{333333} Broken\_end}                            & {\color[HTML]{333333} 0.712±0.069}          & {\color[HTML]{333333} 0.441±0.111}   & {\color[HTML]{333333} 0.708±0.094}          & {\color[HTML]{333333} \textbf{0.778±0.068}}     & {\color[HTML]{333333} {\ul 0.724±0.052}}    & {\color[HTML]{333333} 0.658±0.111}          & {\color[HTML]{333333} {\ul 0.732±0.065}}    & {\color[HTML]{333333} 0.693±0.099}          & {\color[HTML]{333333} 0.712±0.068}              & {\color[HTML]{333333} \textbf{0.825±0.020}} \\
{\color[HTML]{333333} }                                      & {\color[HTML]{333333} Broken\_pick}                           & {\color[HTML]{333333} 0.552±0.003}          & {\color[HTML]{333333} 0.476±0.070}   & {\color[HTML]{333333} \textbf{0.731±0.072}} & {\color[HTML]{333333} 0.644±0.039}              & {\color[HTML]{333333} {\ul 0.659±0.047}}    & {\color[HTML]{333333} 0.585±0.028}          & {\color[HTML]{333333} 0.555±0.027}          & {\color[HTML]{333333} \textbf{0.760±0.037}} & {\color[HTML]{333333} 0.629±0.012}              & {\color[HTML]{333333} {\ul 0.675±0.020}}    \\
{\color[HTML]{333333} }                                      & {\color[HTML]{333333} Cut\_selvage}                           & {\color[HTML]{333333} 0.689±0.016}          & {\color[HTML]{333333} 0.434±0.149}   & {\color[HTML]{333333} {\ul 0.739±0.101}}    & {\color[HTML]{333333} 0.681±0.077}              & {\color[HTML]{333333} \textbf{0.790±0.038}} & {\color[HTML]{333333} 0.709±0.039}          & {\color[HTML]{333333} 0.682±0.025}          & {\color[HTML]{333333} {\ul 0.777±0.036}}    & {\color[HTML]{333333} 0.770±0.014}              & {\color[HTML]{333333} \textbf{0.856±0.009}} \\
{\color[HTML]{333333} }                                      & {\color[HTML]{333333} Fuzzyball}                              & {\color[HTML]{333333} 0.617±0.075}          & {\color[HTML]{333333} 0.525±0.157}   & {\color[HTML]{333333} 0.538±0.092}          & {\color[HTML]{333333} {\ul 0.650±0.064}}        & {\color[HTML]{333333} \textbf{0.771±0.021}} & {\color[HTML]{333333} 0.734±0.039}          & {\color[HTML]{333333} 0.677±0.223}          & {\color[HTML]{333333} 0.701±0.093}          & {\color[HTML]{333333} {\ul 0.842±0.026}}        & {\color[HTML]{333333} \textbf{0.875±0.017}} \\
{\color[HTML]{333333} }                                      & {\color[HTML]{333333} Nep}                                    & {\color[HTML]{333333} {\ul 0.722±0.023}}    & {\color[HTML]{333333} 0.517±0.059}   & {\color[HTML]{333333} 0.717±0.052}          & {\color[HTML]{333333} 0.710±0.044}              & {\color[HTML]{333333} \textbf{0.731±0.019}} & {\color[HTML]{333333} {\ul 0.810±0.042}}    & {\color[HTML]{333333} 0.740±0.052}          & {\color[HTML]{333333} 0.750±0.038}          & {\color[HTML]{333333} 0.771±0.032}              & {\color[HTML]{333333} \textbf{0.908±0.008}} \\
{\color[HTML]{333333} }                                      & {\color[HTML]{333333} Weft\_crack}                            & {\color[HTML]{333333} 0.586±0.134}          & {\color[HTML]{333333} 0.400±0.029}   & {\color[HTML]{333333} {\ul 0.669±0.045}}    & {\color[HTML]{333333} 0.582±0.108}              & {\color[HTML]{333333} \textbf{0.686±0.013}} & {\color[HTML]{333333} 0.599±0.137}          & {\color[HTML]{333333} 0.370±0.037}          & {\color[HTML]{333333} \textbf{0.717±0.072}} & {\color[HTML]{333333} 0.618±0.172}              & {\color[HTML]{333333} {\ul 0.697±0.014}}    \\ \cmidrule(l){2-12} 
\multirow{-7}{*}{{\color[HTML]{333333} \textbf{AITEX}}}      & {\color[HTML]{333333} \textbf{Mean}}                          & {\color[HTML]{333333} 0.646±0.034}          & {\color[HTML]{333333} 0.466±0.030}   & {\color[HTML]{333333} {\ul 0.684±0.033}}    & {\color[HTML]{333333} 0.674±0.034}              & {\color[HTML]{333333} \textbf{0.727±0.057}} & {\color[HTML]{333333} 0.683±0.032}          & {\color[HTML]{333333} 0.626±0.041}          & {\color[HTML]{333333} {\ul 0.733±0.009}}    & {\color[HTML]{333333} 0.724±0.032}              & {\color[HTML]{333333} \textbf{0.806±0.090}} \\ \bottomrule
\end{tabular}}
\label{tab:hard}
\end{table*}

\subsection{Results under General Setting} 

The comparison results of different methods under the general setting are shown in Table~\ref{tab:general}. 

\textbf{Ten Training Anomaly Samples.} Although the six datasets are derived from different scenarios, our approach achieves the best average AUC performance on five of them and the second-best results on MastCam. The upper block in Table~\ref{tab:general} shows the performance at the level of data subsets, where our method outperforms other competitors in most cases.  
Our method achieves a boost of up to 4.4\% compared to the second-best detector on Toothbrush.

\textbf{One Training Anomaly Sample.} By reducing the number of training anomaly samples from ten to one, the performance degradation of our method is significantly less than that of the other competitors.  
Our method obtains the best AUC results on all six datasets, with a maximum improvement of 21.9\% compared to the second-best method on Capsule of MVTec AD. Meanwhile, our method achieves the smallest average AUC performance drop, with 11.4\%, which is much better than DevNet (22.2\%), MLEP (20.8\%), SAOE (14.9\%) and DRA (13.5\%).
On MVTec AD and Optical datasets, our average results trained with one anomaly sample significantly surpass some competitive methods trained with ten anomaly samples. 
In the more detailed results for data subsets of MVTec AD, our results trained with one anomaly outperforms all competitors trained with ten anomalies on Carpet, Grid, and Pill1.
As the number of training anomaly samples decreases, our method suffers the least average AUC results reduction compared to the competing methods, evidencing the better generalization ability of our method.

\subsection{Results under Hard Setting}

Table~\ref{tab:hard} presents the comparison results on the five datasets under the hard setting.

\textbf{Ten Training Anomaly Samples.} Our method performs the best on all five datasets, improving mean AUC results by at least 0.6\% compared to the best contestant. The maximum performance gain of 7.3\% is achieved on AITEX, verifying the effectiveness of our method.
Analysis at the data subsets level shows that our approach also achieves the best or the second-best detection performance in most cases.

\textbf{One Training Anomaly Sample.} Under the condition of one training anomaly sample, our approach achieves the best performance on Carpet, Metal\_nut, AITEX, and ELPV, with the second-best results on MastCam. In regard to the mean performance on MastCam, the results of our methods are comparable to the ensemble method SAOE.
Compared to the best contender, mean AUC performance increases by 2.4\%-5.9\%. 
Compared to the scenarios with ten training anomaly samples, our method reduces AUC performance by an average of 3.68\% on the five datasets, which is better than DRA (3.74\%), MLEP (9.80\%) and DevNet (7.16\%), and close to SAOE (3.50\%).
Moreover, our model trained with one anomaly significantly exceeds competitive methods with ten anomalies on many of the data subsets as well as the datasets, such as DevNet, MLEP, DRA and SAOE on Carpet, its data subsets and Cur\_selvage of AITEX.

\subsection{Results under Anomaly-Free Setting}

Table~\ref{tab:anomalyfree} presents the comparison results on the five texture datasets of MVTec AD under the anomaly-free scenario.
Although there are no training anomaly samples, our approach can still work and shows better detection performance compared to the best recent anomaly-free detection methods, with an average AUC results improvement of 1\% compared to the second best.
This experiment demonstrates the effectiveness of SaliencyCut, which generates pseudo anomalies to promote the learning of unseen anomalies.
At the dataset level, our method significantly outperforms other detectors on Carpet, Tile, and Wood, with the following best performance on Grid.

Compared to the competing methods under the general setting shown in Table~\ref{tab:general}, our model trained without anomaly samples is even better than other detectors training with ten anomaly samples, such as DevNet, MLEP, SAOE and DRA on Carpet, and DevNet and SAOE on Grid.
This justifies the significantly better generalization ability of our method  in detecting unseen anomaly samples.

\begin{table}[]
\caption{AUC results under the anomaly-free setting, all models are trained without anomaly samples.}
\centering
\resizebox{0.79\linewidth}{!}{
\begin{tabular}{@{}c|cccc@{}}
\toprule
\textbf{Dataset} & \textbf{KDAD} & \textbf{SSM}   & \textbf{CutPaste}    & \textbf{Ours}        \\ \midrule
Carpet           & 0.792±0.008   & 0.763          & {\ul 0.939}       & \textbf{0.984±0.011} \\
Grid             & 0.780±0.006   & \textbf{1.000} & \textbf{1.000}    & {\ul 0.974±0.012}    \\
Leather          & 0.950±0.002   & {\ul 0.999}    & \textbf{1.000}    & 0.989±0.002          \\
Tile             & 0.915±0.006   & 0.944          & {\ul 0.946}       & \textbf{0.981±0.008} \\
Wood             & 0.942±0.002   & 0.959          & {\ul 0.991}       & \textbf{0.997±0.002} \\ \midrule
\textbf{Mean}    & 0.875±0.074   & 0.933±0.087    & {\ul 0.975±0.026} & \textbf{0.985±0.011} \\ \bottomrule
\end{tabular}}
\label{tab:anomalyfree}
\end{table}

\begin{table}[]
\caption{Ablation study of our method and its variants.}
\centering
\resizebox{\linewidth}{!}{
\begin{tabular}{@{}cc|ccccc@{}}
\toprule
\multicolumn{2}{c|}{{\color[HTML]{333333} Four-head learning}} & {\color[HTML]{333333} \textbf{\checkmark}} & {\color[HTML]{333333} \textbf{-}} & -                 & {\color[HTML]{333333} \textbf{-}} & -                    \\
\multicolumn{2}{c|}{Two-head learning}                         & -                                 & \checkmark                                 & \checkmark                 & \checkmark                                 & \checkmark                    \\
\multicolumn{2}{c|}{Latent residual}                           & \checkmark                                 & \checkmark                                 & \checkmark                 & -                                 & -                    \\
\multicolumn{2}{c|}{Patch-wise residual}                       & -                                 & -                                 & -                 & \checkmark                                 & \checkmark                    \\
\multicolumn{2}{c|}{CutMix}                                    & \checkmark                                 & \checkmark                                 & -                 & \checkmark                                 & -                    \\
\multicolumn{2}{c|}{SaliencyCut}                               & -                                 & -                                 & \checkmark                 & -                                 & \checkmark                    \\ \midrule
                                           & Color             & 0.886±0.042                       & 0.939±0.003                       & {\ul 0.944±0.007} & 0.931±0.010                       & \textbf{0.951±0.005} \\
                                           & Cut               & 0.922±0.038                       & 0.940±0.005                       & 0.955±0.002       & {\ul 0.961±0.006}                 & \textbf{0.964±0.008} \\
                                           & Hole              & 0.947±0.016                       & 0.937±0.014                       & {\ul 0.958±0.010} & 0.949±0.009                       & \textbf{0.961±0.004} \\
                                           & Metal             & 0.933±0.022                       & 0.938±0.008                       & {\ul 0.948±0.006} & {\ul 0.948±0.016}                 & \textbf{0.951±0.002} \\
                                           & Thread            & 0.989±0.004                       & 0.989±0.004                       & {\ul 0.996±0.002} & 0.990±0.001                       & \textbf{0.998±0.000} \\ \cmidrule(l){2-7} 
\multirow{-6}{*}{\textbf{Carpet}}          & \textbf{Mean}     & 0.935±0.013                       & 0.949±0.022                       & {\ul 0.960±0.020} & 0.956±0.022                       & \textbf{0.965±0.018} \\ \midrule
                                           & Bent              & 0.990±0.003                       & 0.994±0.005                       & 0.994±0.002       & {\ul 0.995±0.005}                 & \textbf{0.996±0.003} \\
                                           & Color             & {\ul 0.967±0.011}                 & 0.948±0.031                       & 0.957±0.006       & 0.950±0.005                       & \textbf{0.979±0.004} \\
                                           & Flip              & 0.913±0.021                       & 0.944±0.005                       & 0.901±0.018       & \textbf{0.951±0.009}              & {\ul 0.950±0.013}    \\
                                           & Scratch           & 0.911±0.034                       & 0.967±0.012                       & 0.974±0.004       & {\ul 0.975±0.009}                 & \textbf{0.983±0.002} \\ \cmidrule(l){2-7} 
\multirow{-5}{*}{\textbf{Metal\_nut}}      & \textbf{Mean}     & 0.945±0.017                       & 0.963±0.026                       & 0.957±0.036       & {\ul 0.968±0.020}                 & \textbf{0.977±0.018} \\ \bottomrule
\end{tabular}}
\label{tab:ablation}
\end{table}

\subsection{Ablation Studies}
\label{exp:ablation}

Table~\ref{tab:ablation} shows ablation study results under the hard setting on Carpet and Metal\_nut with ten training anomaly samples.
We select DRA~\cite{ding2022catching} as the baseline built with one four-head learning mechanism, one latent residual module, and CutMix augmentation.
Compared to utilizing CutMix augmentation, SaliencyCut improves performance on two datasets by 0.9\% on average under two-head learning and patch-wise residual paradigm, indicating its efficiency in detecting unseen anomaly samples.

Our two-head learning mechanism outperforms the four-head learning mechanism on various data subsets, particularly on Color of Carpet and Scratch of Metal\_nut, which demonstrates the effectiveness of the former approach and validates our theory. Replacing the latent residual module with the patch-wise residual module leads to further performance improvement on almost all data subsets and datasets. 
It verifies that the patch-wise residual allows the network to learn the discriminative features of anomalies.

\subsection{Sensitivity Analysis}
\label{exp:sensitive}

\begin{figure}
\centering
\includegraphics[scale=0.25]{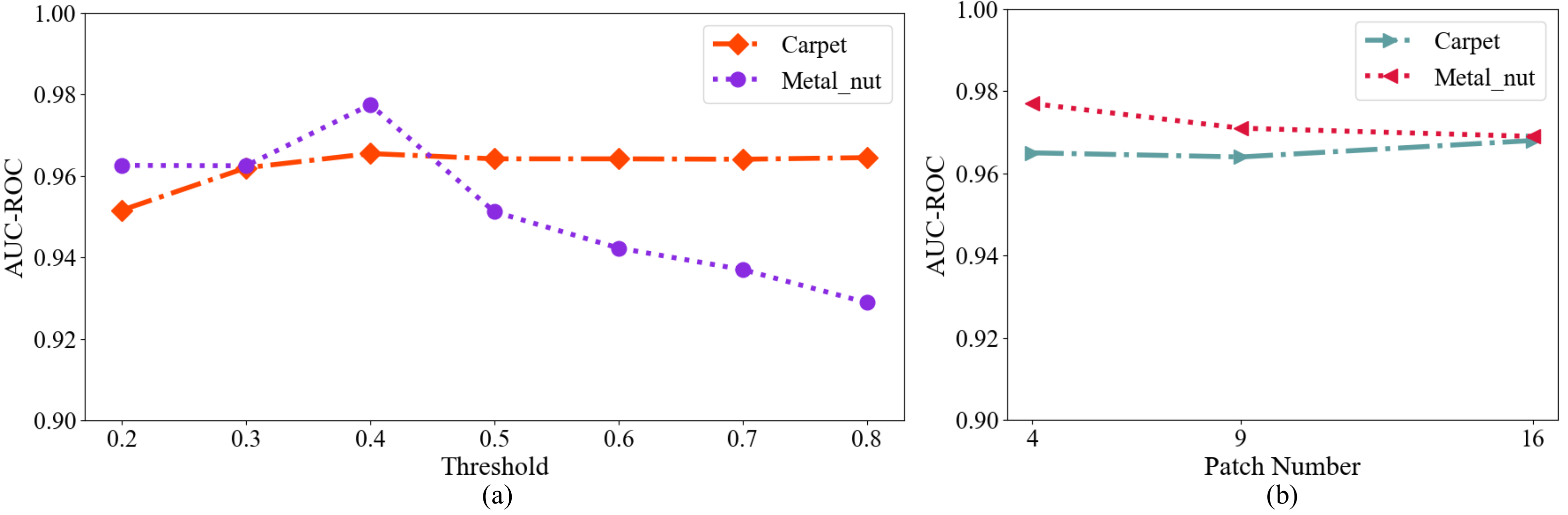}
\caption{Sensitivity analysis of threshold (left) and patch number (right) on Carpet and Metal\_nut under the hard setting.}
\label{fig:sensitive}
\end{figure}

\textbf{Threshold for Obtaining Saliency Masks.} Figure~\ref{fig:sensitive}\textcolor{red}{a} shows AUC results with varying thresholds on Carpet and Metal\_nut under the hard setting.
We successively set the threshold from 0.2 to 0.8 with a step size of 0.1 to analyse the sensitivity.
When the threshold is set to 0.2, our method is able to generate effective pseudo-anomaly samples.
As the threshold increases from 0.2 to 0.4, and the size of pasted non-saliency regions becomes smaller, our model obtains the best detection performance.
With the increase of threshold, the size of non-saliency regions decreases, and the effectiveness of generated pseudo anomaly samples drops.
Thus, the threshold is recommended to be set to 0.4 for producing suitable pseudo anomaly samples.

\textbf{Patch Number.} Figure~\ref{fig:sensitive}\textcolor{red}{b} shows AUC results with varying patch number on Carpet and Metal\_nut under the hard setting. Our model shows less sensitivity to patch numbers, as all sizes deliver good performance. 

\section{Conclusion}

We design a two-head learning mechanism with a patch-wise residual learning block, and a novel augmentation method SaliencyCut guided by gradients to detect fine-grained anomalies in open-set scenarios. 
Extensive experiments under both general and hard settings evidence that our method overall outperforms the competing methods.
Additionally, our method can work under the anomaly-free setting and beat three anomaly-free detection models.

{\small
\bibliographystyle{ieee_fullname}
\bibliography{egbib}
}

\end{document}